\RequirePackage{fix-cm}
\documentclass[final,twocolumn,natbib]{svjour3}
\usepackage{graphicx}
\usepackage{multirow}
\usepackage[figuresright]{rotating}
\usepackage{amsmath}
\usepackage{natbib}
\usepackage[table, svgnames, dvipsnames]{xcolor}
\usepackage{hyperref}
\smartqed  % flush right qed marks, e.g. at end of proof
\begin{document}
\title{Detecting Anomalies within Smart Buildings using Do-It-Yourself Internet of Things}
% \subtitle{Do you have a subtitle?\\ If so, write it here}


\author{Yasar Majib         \and
        Mahmoud Barhamgi \and
        Behzad Momahed Heravi \and
        Sharadha Kariyawasam \and
        Charith Perera
}
\institute{Y. Majib \at
            Cardiff University, UK. 
              \email{MajibY@Cardiff.ac.uk}
\date{Received: date / Accepted: date}}
\maketitle
\begin{abstract}
Detecting anomalies at the time of happening is vital in environments like buildings and homes to identify potential cyber-attacks. This paper discussed the various mechanisms to detect anomalies as soon as they occur. We shed light on crucial considerations when building machine learning models. We constructed and gathered data from multiple self-build (DIY) IoT devices with different in-situ sensors and found effective ways to find the point, contextual and combine anomalies. We also discussed several challenges and potential solutions when dealing with sensing devices that produce data at different sampling rates and how we need to pre-process them in machine learning models. This paper also looks at the pros and cons of extracting sub-datasets based on environmental conditions.
\keywords{Anomaly Detection \and Machine Learning \and Internet of Things \and Smart Buildings}
\end{abstract}
%%%%%%%%%%%%%%%%%%%%%%%%%%%%%%
% Introduction content removal
%  As D. Hawkins \cite{Hawkins1980} defined it "an observation which deviates so significantly from other observations as to arouse suspicion that it was generated by a different mechanism". 
%%%%%%%%%%%%%%%%%%%%%%%%%%%%%%

\section{Introduction}
An anomaly is something unexpected, abnormal or distanced from the ordinary. From a technology perspective, an anomaly results from equipment malfunction, cyber or physical intrusion, financial fraud (e.g. credit card usage by hackers), terrorist activity, and an abrupt change detected by sensors in the physical environment due to an accident. Following are the types of anomalies: 
\begin{enumerate}
    \item Point Anomalies: A single sample, different from normal samples. For example, a credit card (CC) transaction with an amount much larger than the CC holder's routine transactions.
    \item Collective Anomalies: A sample is a collection of several data points considered anomalous if it differs from other samples. For example, an electrocardiogram (ECG) is a collection of readings of the heart's activity over a specific period as one data sample.
    \item Contextual Anomalies: If a sample is contextually different from normal samples. Time is the context in time-series data considering a situation where data is streaming from sensors. An anomalous sample depends on a set of time-series values, e.g. a temperature trend of the last 30 minutes showing 20\textsuperscript{o}C increases 50\% abruptly. In some other time (context) 30\textsuperscript{o}C is considered normal temperature.
\end{enumerate}
Our work looked into all the above types of anomalies in our dataset. We proposed multiple solutions to look for abnormalities in various contexts, e.g. time-series, multivariate, and inter-device sensor combinations. The high-level idea behind anomaly detection is to i) save resources by finding faults in systems in advance, ii) respond to events as early as possible iii) deal with security breaches. Equipment with the least latency from sensors is microcontrollers, and these devices are resource-constrained. With the rapidly growing IoT domain, there are a few off-the-shelf microcontrollers available now \cite{Sudharsan2021} which support machine learning (ML) on edge using libraries, e.g. TensorFlow. Detecting anomalies as soon as they occur can help save a building from various challenges. Gas leakage by equipment malfunction or pipeline cracks, discomfort due to a sudden change in environment (temperature, humidity, noise, air quality, and others), infrastructure damage, physical access at a non-working time, or unauthorised personnel cyber-physical attacks related. Detecting anomalies at the edge ensures early response and reduces the risk of it getting ignored by the central system in case of unavailability of network connectivity due to technical problems or cyber-attacks, e.g. Daniel of Service (DoS).
We collected data from self-built physical devices with 32 data streams from 14 unique sensors. We have combined intra-device data streams and inter-devices unique sensors' streams. Other than the original "unconditional" dataset, we applied two (02) environmental conditions to the data set, then applied data preprocessing (scaling and reduction) techniques to each resulting data set and then used different ML algorithms. We tested all models using both normal and anomaly data sets and presented the results in HTML format at  \href{https://gitlab.com/IOTGarage/cyphyradar}{GitLab/CyPhyRadar}. We evaluated the models based on computational time vs the number of detected anomalies.
\subsection{Contributions}
\begin{itemize}
    \item Impact of environmental conditions' based data set in anomaly detection
    \item Pros and cons of conventional (scaling/reduction) and unconventional (atan) data preprocessing methods
    \item Comparison of different ML techniques
    \item Relations between various sensors in the context of discovering anomalies in building
    \item Best practices to transform univariate data into time-series format
    \item Handling missing data and synchronizing data streams from different devices
\end{itemize}
\section{Anomaly Detection within Smart Buildings}
It is not energy-saving anymore; it is about the overall resilience of smart buildings, which is the next big challenge. Smart buildings require mechanisms to mitigate or prevent fire, gas leakages, attacks, disasters, accidents, safety and security-related issues, and other unforeseen challenges. Secondary sensor networks can help mitigate such events by observing physical channels such as external eyes and ears. Any compromise-able device in a cyber network can allow attackers to gain control over the complete building management systems \cite{AlexSchiffer2017}.
\subsection{Data Collection Setup}
\begin{figure*}
    \includegraphics[scale=.58]{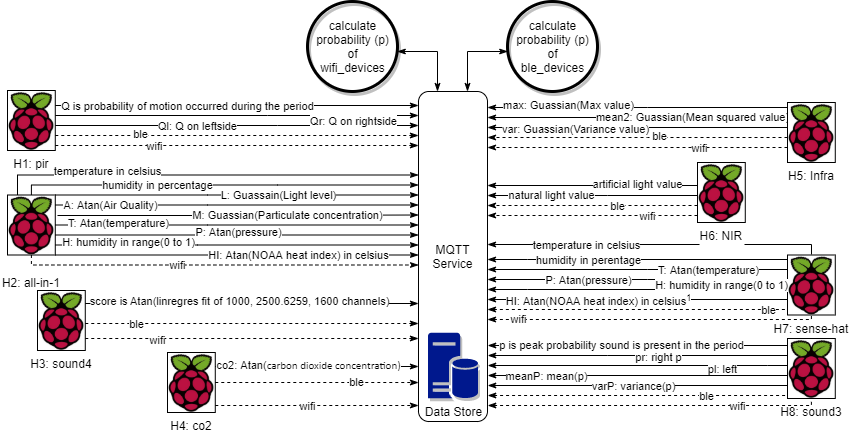}
    \caption{H1: Passive InfraRed, H2: All-in-1 Multi-sensor, H3: Sound4, H4: Carbon Dioxide, H5: Infra-sound, H6: Light, H7: Sense-Hat Multi-sensor, H8: Sound3}
    \label{fig:SensorNetworkDiagram}
\end{figure*}

We have implemented a sensing network consisting of various 14 different environmental sensors, Arduino based microcontrollers and RaspberryPi (RPi) microprocessors, as shown in Table \ref{tab:DataStreamsDetails}. The sensor reads the environmental changes and transfers readings to the attached RPi, directly or through a microcontroller, which then transforms and/or transfers these values to the ingestor using unique Message Queuing Telemetry Transport (MQTT) channels. The data set consists of 32 different data streams from eight (8) device sets, i.e. sensor-Arduino-RPi (DSet). Temperature, humidity sensors and some other associated data streams were duplicated in two device-sets; although both device-sets were at the same place, one of the DSet's sensors was influenced by a nearby heat source. Thus, the readings are different in these data streams. Timestamp and other properties were added to every new entry by the ingestor before inserting it into the data set. The probability of BLE and WiFi devices in the area was also calculator by the ingestor after receiving collective BLE and WiFi devices' information from all other physical devices; these data streams in channels ble\_devices and wifi\_devices were considered as virtual devices.  Figure \ref{fig:SensorNetworkDiagram} shows the overall architecture of data collection setup, processing points, devices' and channels' names. We divided the data sets from July 24, 2020, to January 7, 2021, and from March 26, 2021, to July 16, 2021, into two subsets, normal and abnormal, respectively. Both data sets were captured during normal routine operations, and some naturally occurring unusual activities were recorded in the time-frames of both data sets. We used the normal subset for training and testing machine learning models, whereas the anomaly subset was for testing purposes only. 
\begin{itemize}
    \item Physical Devices = 8
    \item Virtual Devices = 2
    \item Environmental Conditions = 3
    \item Pre-processing Techniques = 8
    \item Data Streams (Total) = 32
    \item Intra-Device Combinations = 626
    \item Data Streams (Unique Sensors) = 14
    \item Inter-Device Combinations of unique sensors = 16383
    \item Machine Learning Techniques = 4
\end{itemize}
\subsection{Data Collection Challenges}
Some of the main challenges in data collection are:
\begin{itemize}
    \item Time synchronisation, microcontrollers do not come with an internal time clock, making it tricky to keep data synchronised from different host devices, assuming the reporting time between each device is different.
    \item Handling heterogeneous data types, contexts and formats
    \item Low-resolution sensors, e.g., some generate integer values for reading instead of floating-point values, e.g., temperature value 22 instead of 22.0-22.9.
    \item Some sensors generate arbitrary data, which is very difficult to detect and troubleshoot on edge.
    \item Dual channel sensors like temperature-humidity have sensing errors in either of the channels creating difficulty to troubleshoot on edge.
    \item Different communication mediums have different latency, which is also a challenge in time synchronisation.
    \item Communication modules provide limited access to the chip via AT Commands.
    \item Skipped or missed part of data at random times due to equipment malfunction, network connectivity, electric power or other issues.
\end{itemize}
\subsection{Data Cleaning and Normalisation}
\label{DataCleaningandPreProcessing}
We pre-processed the data sets before performing ML-associated operations to save time and computational resources. There were various possible combinations of errors in data sets like null, non-numeric, or irrelevant values when capturing data due to sensor malfunctions or ingestion processing. We removed all rows with null values, converted the date and time into a DataFrame supported format, changed the type data type of all other values to integer or float, and normalised data sets.
\subsection{Data Streams Overview and Analysis}
Analysing all data streams, individually and jointly, is very important before applying operations. Analysis helps in getting a better understanding of data streams and helps in estimating which pre-processing technique with which type of model should be used to do further processing. The best way to visualise data streams is by graphs; we used interact-able graphs using Plotly-library to better understand the data streams from all sensors. We joined data streams from all devices to better understand the relations between each combination. Moreover, the Table \ref{tab:DataStreamsDetails} hosts details of all individual data streams with description, host device, MQTT topic, edge-processing technique (Process), minimum value, maximum value, average, standard deviation (SD), and median absolute deviation (MAD).\\
\begin{sidewaystable}
\begin{center}
\begin{minipage}{\textheight}
\caption{Data Stream Details}
\label{tab:DataStreamsDetails}
\scalebox{0.66}{
\begin{tabular}{lllllllllllllll}
\multicolumn{5}{c}{Data   Stream Properties} & \multicolumn{5}{c}{Normal Data} & \multicolumn{5}{c}{Anomaly Data} \\
Name & Description & Sensor Device & Topic & Process & Min & Max & Average & SD & MAD & Min & Max & Average & SD & MAD \\
\textbf{temp} & \textbf{temperature sensor} & \textbf{Sense-HAT} & \textbf{sense-hat} & None & \cellcolor[HTML]{F1E784}\textbf{3.1} & \cellcolor[HTML]{B9D780}\textbf{26.7} & \cellcolor[HTML]{A4D17F}\textbf{22.5823} & \cellcolor[HTML]{81C77D}\textbf{2.16122} & \cellcolor[HTML]{89C97E}\textbf{1.92738} & \cellcolor[HTML]{A3D17F}\textbf{18.1} & \cellcolor[HTML]{96CD7E}\textbf{30} & \cellcolor[HTML]{A5D17F}\textbf{21.88942} & \cellcolor[HTML]{B6D680}\textbf{1.49556} & \cellcolor[HTML]{AED480}\textbf{1.63086} \\
\textbf{humidity} & \textbf{humidity sensor} & \textbf{Sense-HAT} & \textbf{sense-hat} & None & \cellcolor[HTML]{63BE7B}\textbf{32.4} & \cellcolor[HTML]{63BE7B}\textbf{58.3} & \cellcolor[HTML]{63BE7B}\textbf{38.4451} & \cellcolor[HTML]{92CC7E}\textbf{1.88704} & \cellcolor[HTML]{7FC77D}\textbf{2.07564} & \cellcolor[HTML]{63BE7B}\textbf{30.5} & \cellcolor[HTML]{63BE7B}\textbf{43.8} & \cellcolor[HTML]{63BE7B}\textbf{37.35104} & \cellcolor[HTML]{91CC7E}\textbf{2.1766} & \cellcolor[HTML]{89C97E}\textbf{2.37216} \\
T & Atan transformed temperature & Sense-HAT & sense-hat & Atan & \cellcolor[HTML]{FFEB84}0.01965 & \cellcolor[HTML]{FEE683}0.94034 & \cellcolor[HTML]{FEEB84}0.72561 & \cellcolor[HTML]{F9EA84}0.25949 & \cellcolor[HTML]{FBEA84}0.09736 & \cellcolor[HTML]{FFEB84}0.11915 & \cellcolor[HTML]{FFEB84}0.96171 & \cellcolor[HTML]{FEEB84}0.636083 & \cellcolor[HTML]{FBEA84}0.22574 & \cellcolor[HTML]{F2E884}0.29127 \\
\textbf{P} & \textbf{Atan transformed pressure} & \textbf{Sense-HAT} & \textbf{sense-hat} & Atan & \cellcolor[HTML]{FFEB84}\textbf{0.13972} & \cellcolor[HTML]{FEE482}\textbf{0.93009} & \cellcolor[HTML]{FDEB84}\textbf{0.83703} & \cellcolor[HTML]{FDD47F}\textbf{0.13184} & \cellcolor[HTML]{FFEB84}\textbf{0.03382} & \cellcolor[HTML]{FDEB84}\textbf{0.5869} & \cellcolor[HTML]{FEE783}\textbf{0.93243} & \cellcolor[HTML]{FDEB84}\textbf{0.878565} & \cellcolor[HTML]{FA9473}\textbf{0.05539} & \cellcolor[HTML]{FFEB84}\textbf{0.02667} \\
H & humidity scaled 0-1 & Sense-HAT & sense-hat & Scale & \cellcolor[HTML]{FEEB84}0.324 & \cellcolor[HTML]{FBA877}0.583 & \cellcolor[HTML]{FFEB84}0.38445 & \cellcolor[HTML]{F86E6C}0.01887 & \cellcolor[HTML]{FFEB84}0.02076 & \cellcolor[HTML]{FEEB84}0.305 & \cellcolor[HTML]{FA9373}0.438 & \cellcolor[HTML]{FFEB84}0.37351 & \cellcolor[HTML]{F8736D}0.02177 & \cellcolor[HTML]{FEEA83}0.02372 \\
HI & Atan transformed NOAA heat index & Sense-HAT & sense-hat & Atan & \cellcolor[HTML]{FFEB84}0.01757 & \cellcolor[HTML]{FEE582}0.93836 & \cellcolor[HTML]{FEEB84}0.65749 & \cellcolor[HTML]{F8E984}0.28259 & \cellcolor[HTML]{F7E984}0.16325 & \cellcolor[HTML]{FFEB84}0.08645 & \cellcolor[HTML]{FFEB84}0.96135 & \cellcolor[HTML]{FFEB84}0.511415 & \cellcolor[HTML]{F8E984}0.27952 & \cellcolor[HTML]{EEE683}0.37915 \\
\textbf{max} & \textbf{Guassian transformed maximum   value} & \textbf{Infiltec Model INFRA20} & \textbf{infra} & Guassain & \cellcolor[HTML]{F8696B}\textbf{4E-10} & \cellcolor[HTML]{FFEB84}\textbf{0.9966} & \cellcolor[HTML]{F8696B}\textbf{0.00338} & \cellcolor[HTML]{F97D6E}\textbf{0.03547} & \cellcolor[HTML]{F8696B}\textbf{2.8E-06} & \cellcolor[HTML]{F8696B}\textbf{4E-10} & \cellcolor[HTML]{FEDF81}\textbf{0.88429} & \cellcolor[HTML]{F8696B}\textbf{0.002099} & \cellcolor[HTML]{F8716C}\textbf{0.02009} & \cellcolor[HTML]{F8696B}\textbf{8.7E-07} \\
mean2 & Guassian transformed mean   squared value & Infiltec Model INFRA20 & infra & Atan & \cellcolor[HTML]{F8696B}7.8E-12 & \cellcolor[HTML]{FFEB84}0.99999 & \cellcolor[HTML]{F8696B}0.00433 & \cellcolor[HTML]{F9816F}0.03985 & \cellcolor[HTML]{F8696B}5.4E-06 & \cellcolor[HTML]{F8696B}5.8E-12 & \cellcolor[HTML]{FEE482}0.9163 & \cellcolor[HTML]{F8696B}0.002814 & \cellcolor[HTML]{F8766D}0.0249 & \cellcolor[HTML]{F8696B}2.1E-06 \\
var & Guassian transformed variance   value & Infiltec Model INFRA20 & infra & None & \cellcolor[HTML]{F8696B}1.8E-05 & \cellcolor[HTML]{FFEB84}0.99859 & \cellcolor[HTML]{F8786D}0.03748 & \cellcolor[HTML]{FCB87A}0.10021 & \cellcolor[HTML]{F97D6F}0.00334 & \cellcolor[HTML]{F8696B}1.9E-05 & \cellcolor[HTML]{FFEB84}0.99997 & \cellcolor[HTML]{F8786D}0.037592 & \cellcolor[HTML]{FCBA7A}0.09251 & \cellcolor[HTML]{F8766D}0.00238 \\
\textbf{C} & \textbf{Atan transformed CO2   concentration} & \textbf{DFRobot SEN0219} & \textbf{co2} & None & \cellcolor[HTML]{FFEB84}\textbf{0.15119} & \cellcolor[HTML]{FDD880}\textbf{0.85903} & \cellcolor[HTML]{FDD780}\textbf{0.2504} & \cellcolor[HTML]{FBB179}\textbf{0.09349} & \cellcolor[HTML]{FFEB84}\textbf{0.0339} & \cellcolor[HTML]{FFEB84}\textbf{0.18182} & \cellcolor[HTML]{FDD57F}\textbf{0.82685} & \cellcolor[HTML]{FEDE81}\textbf{0.27269} & \cellcolor[HTML]{FCBF7B}\textbf{0.09757} & \cellcolor[HTML]{FEEB84}\textbf{0.04537} \\
\textbf{natural} & \textbf{Probability of natural light} & \textbf{SparkFun AS7263} & \textbf{nir} & None & \cellcolor[HTML]{F8696B}\textbf{0} & \cellcolor[HTML]{FEE582}\textbf{0.9382} & \cellcolor[HTML]{F8796E}\textbf{0.03925} & \cellcolor[HTML]{FEDD81}\textbf{0.14193} & \cellcolor[HTML]{F8696B}\textbf{0} & \cellcolor[HTML]{F8696B}\textbf{0} & \cellcolor[HTML]{FEDE81}\textbf{0.88205} & \cellcolor[HTML]{F98570}\textbf{0.068757} & \cellcolor[HTML]{FDEB84}\textbf{0.18547} & \cellcolor[HTML]{F8696B}\textbf{0} \\
\textbf{artificial} & \textbf{Probability of artificial light} & \textbf{SparkFun AS7263} & \textbf{nir} & None & \cellcolor[HTML]{F8696B}\textbf{0} & \cellcolor[HTML]{FFEB84}\textbf{0.98599} & \cellcolor[HTML]{FFEB84}\textbf{0.33197} & \cellcolor[HTML]{EDE683}\textbf{0.4551} & \cellcolor[HTML]{F8696B}\textbf{0} & \cellcolor[HTML]{F8696B}\textbf{0} & \cellcolor[HTML]{FFEB84}\textbf{0.99736} & \cellcolor[HTML]{FFEB84}\textbf{0.324322} & \cellcolor[HTML]{EFE784}\textbf{0.45155} & \cellcolor[HTML]{F8696B}\textbf{0} \\
\textbf{p} & \textbf{Peak probability of presense of   sound} & \textbf{NA} & \textbf{sound3} & None & \cellcolor[HTML]{F8696B}\textbf{0} & \cellcolor[HTML]{FFEB84}\textbf{1} & \cellcolor[HTML]{FDD47F}\textbf{0.24323} & \cellcolor[HTML]{F1E784}\textbf{0.39295} & \cellcolor[HTML]{FEEA83}\textbf{0.02058} & \cellcolor[HTML]{F8696B}\textbf{0} & \cellcolor[HTML]{FFEB84}\textbf{1} & \cellcolor[HTML]{FED980}\textbf{0.262058} & \cellcolor[HTML]{F1E784}\textbf{0.40104} & \cellcolor[HTML]{FFEB84}\textbf{0.02377} \\
pr & Peak probability of presense of   sound on right side & NA & sound3 & None & \cellcolor[HTML]{F8696B}0 & \cellcolor[HTML]{FFEB84}1 & \cellcolor[HTML]{FEDA80}0.25626 & \cellcolor[HTML]{F2E884}0.3752 & \cellcolor[HTML]{FFEB84}0.03195 & \cellcolor[HTML]{F8696B}0 & \cellcolor[HTML]{FFEB84}1 & \cellcolor[HTML]{FEE082}0.278179 & \cellcolor[HTML]{F3E884}0.37982 & \cellcolor[HTML]{FFEB84}0.03816 \\
pl & Peak probability of presense of   sound on left side & NA & sound3 & None & \cellcolor[HTML]{F8696B}0 & \cellcolor[HTML]{FFEB84}1 & \cellcolor[HTML]{FDD47F}0.24323 & \cellcolor[HTML]{F1E784}0.39295 & \cellcolor[HTML]{FEEA83}0.02058 & \cellcolor[HTML]{F8696B}0 & \cellcolor[HTML]{FFEB84}1 & \cellcolor[HTML]{FED980}0.262058 & \cellcolor[HTML]{F1E784}0.40104 & \cellcolor[HTML]{FFEB84}0.02377 \\
meanP & Mean probability of presense of   sound & NA & sound3 & None & \cellcolor[HTML]{F8696B}0 & \cellcolor[HTML]{FFEB84}1 & \cellcolor[HTML]{FBAE78}0.15991 & \cellcolor[HTML]{F4E884}0.33798 & \cellcolor[HTML]{F86A6B}0.00017 & \cellcolor[HTML]{F8696B}0 & \cellcolor[HTML]{FFEB84}1 & \cellcolor[HTML]{FBAD78}0.158854 & \cellcolor[HTML]{F5E884}0.33858 & \cellcolor[HTML]{F86A6B}0.00032 \\
varP & Variance of probability of   presense of sound & NA & sound3 & None & \cellcolor[HTML]{F8696B}0 & \cellcolor[HTML]{FFEB84}1 & \cellcolor[HTML]{FBB379}0.1694 & \cellcolor[HTML]{F2E884}0.36446 & \cellcolor[HTML]{F8696B}1.2E-14 & \cellcolor[HTML]{F8696B}0 & \cellcolor[HTML]{FFEB84}1 & \cellcolor[HTML]{FCB579}0.179248 & \cellcolor[HTML]{F3E884}0.37252 & \cellcolor[HTML]{F8696B}7.8E-14 \\
\textbf{score} & \textbf{Atan transformed of linregres   fit of 1.6K channels} & \textbf{NA} & \textbf{sound4} & None & \cellcolor[HTML]{FEEB84}\textbf{0.24367} & \cellcolor[HTML]{FBAC77}\textbf{0.60479} & \cellcolor[HTML]{FFEB84}\textbf{0.51678} & \cellcolor[HTML]{F8786D}\textbf{0.02923} & \cellcolor[HTML]{FA9D75}\textbf{0.00835} & \cellcolor[HTML]{FEEB84}\textbf{0.24875} & \cellcolor[HTML]{FBAA77}\textbf{0.57387} & \cellcolor[HTML]{FFEB84}\textbf{0.51733} & \cellcolor[HTML]{F8776D}\textbf{0.02633} & \cellcolor[HTML]{F98470}\textbf{0.00509} \\
\textbf{Q} & \textbf{Probability of motion} & \textbf{Passive InfraRed} & \textbf{pir} & None & \cellcolor[HTML]{FFEB84}\textbf{0.029} & \cellcolor[HTML]{FBA276}\textbf{0.545} & \cellcolor[HTML]{F98470}\textbf{0.06509} & \cellcolor[HTML]{FA9F75}\textbf{0.07245} & \cellcolor[HTML]{F8726C}\textbf{0.00148} & \cellcolor[HTML]{FFEB84}\textbf{0.03} & \cellcolor[HTML]{FBA476}\textbf{0.541} & \cellcolor[HTML]{F98370}\textbf{0.063051} & \cellcolor[HTML]{FA9E75}\textbf{0.06494} & \cellcolor[HTML]{F8716C}\textbf{0.00148} \\
Qr & Probability of motion on right   side & Passive InfraRed & pir & None & \cellcolor[HTML]{FEE883}0.017 & \cellcolor[HTML]{F8736C}0.27 & \cellcolor[HTML]{F8776D}0.03582 & \cellcolor[HTML]{F97F6F}0.03701 & \cellcolor[HTML]{F8726C}0.00148 & \cellcolor[HTML]{FEDF81}0.018 & \cellcolor[HTML]{F87A6E}0.292 & \cellcolor[HTML]{F8766D}0.034289 & \cellcolor[HTML]{F97E6F}0.03283 & \cellcolor[HTML]{F8716C}0.00148 \\
Ql & Probability of motion on left   side & Passive InfraRed & pir & None & \cellcolor[HTML]{FCC37C}0.012 & \cellcolor[HTML]{F8786D}0.3 & \cellcolor[HTML]{F8746D}0.02927 & \cellcolor[HTML]{F97E6F}0.03617 & \cellcolor[HTML]{F8726C}0.00148 & \cellcolor[HTML]{FCB77A}0.012 & \cellcolor[HTML]{F87A6E}0.291 & \cellcolor[HTML]{F8746D}0.028704 & \cellcolor[HTML]{F97E6F}0.03283 & \cellcolor[HTML]{F8716C}0.00148 \\
temperature & temperature sensor & Sense-HAT & all-in-1 & None & \cellcolor[HTML]{DAE182}7.7 & \cellcolor[HTML]{B5D680}28.3 & \cellcolor[HTML]{9CCF7F}24.7235 & \cellcolor[HTML]{A6D27F}1.56751 & \cellcolor[HTML]{A4D17F}1.4826 & \cellcolor[HTML]{D9E082}7.6 & \cellcolor[HTML]{9FD07F}27.4 & \cellcolor[HTML]{99CE7F}24.65208 & \cellcolor[HTML]{C3DA81}1.26133 & \cellcolor[HTML]{B6D680}1.4826 \\
humidity & humiditiy sensor & Sense-HAT & all-in-1 & None & \cellcolor[HTML]{DCE182}7.3 & \cellcolor[HTML]{C9DC81}21 & \cellcolor[HTML]{C9DC81}13.5195 & \cellcolor[HTML]{63BE7B}2.62001 & \cellcolor[HTML]{63BE7B}2.52042 & \cellcolor[HTML]{E4E483}5.3 & \cellcolor[HTML]{BFD981}18.8 & \cellcolor[HTML]{CEDD82}12.10845 & \cellcolor[HTML]{63BE7B}3.01125 & \cellcolor[HTML]{63BE7B}3.11346 \\
\textbf{L} & \textbf{Guassian transformed light level} & \textbf{Sense-HAT} & \textbf{all-in-1} & None & \cellcolor[HTML]{F8696B}\textbf{0} & \cellcolor[HTML]{FFEB84}\textbf{1} & \cellcolor[HTML]{FDCE7E}\textbf{0.23047} & \cellcolor[HTML]{F2E884}\textbf{0.37456} & \cellcolor[HTML]{F86B6B}\textbf{0.00032} & \cellcolor[HTML]{F8696B}\textbf{0} & \cellcolor[HTML]{FFEB84}\textbf{0.99995} & \cellcolor[HTML]{FDCB7E}\textbf{0.229474} & \cellcolor[HTML]{F4E884}\textbf{0.34466} & \cellcolor[HTML]{F87A6E}\textbf{0.0032} \\
\textbf{A} & \textbf{Atan transformed of air quality} & \textbf{Sense-HAT} & \textbf{all-in-1} & None & \cellcolor[HTML]{FCEB84}\textbf{0.64147} & \cellcolor[HTML]{FDCB7D}\textbf{0.78394} & \cellcolor[HTML]{FEEB84}\textbf{0.74921} & \cellcolor[HTML]{F8696B}\textbf{0.01241} & \cellcolor[HTML]{FCB679}\textbf{0.01239} & \cellcolor[HTML]{FCEB84}\textbf{0.65031} & \cellcolor[HTML]{FDD17F}\textbf{0.80128} & \cellcolor[HTML]{FEEB84}\textbf{0.74926} & \cellcolor[HTML]{F8696B}\textbf{0.01134} & \cellcolor[HTML]{FBAA77}\textbf{0.01189} \\
\textbf{M} & \textbf{Guassian transformed of   particulate concentration} & \textbf{Sense-HAT} & \textbf{all-in-1} & None & \cellcolor[HTML]{F8696B}\textbf{0} & \cellcolor[HTML]{FDCE7E}\textbf{0.80237} & \cellcolor[HTML]{F8736D}\textbf{0.02732} & \cellcolor[HTML]{FBA276}\textbf{0.07666} & \cellcolor[HTML]{F98370}\textbf{0.00415} & \cellcolor[HTML]{F8696B}\textbf{0} & \cellcolor[HTML]{FBA777}\textbf{0.55781} & \cellcolor[HTML]{F86F6C}\textbf{0.016666} & \cellcolor[HTML]{F97F6F}\textbf{0.0342} & \cellcolor[HTML]{F98870}\textbf{0.00567} \\
T & Atan transformed temperature & Sense-HAT & all-in-1 & None & \cellcolor[HTML]{FFEB84}0.02646 & \cellcolor[HTML]{FEE883}0.95303 & \cellcolor[HTML]{FDEB84}0.89234 & \cellcolor[HTML]{FA9A74}0.06747 & \cellcolor[HTML]{FFEB84}0.0296 & \cellcolor[HTML]{FFEB84}0.02627 & \cellcolor[HTML]{FEE983}0.94664 & \cellcolor[HTML]{FDEB84}0.900589 & \cellcolor[HTML]{F98670}0.04086 & \cellcolor[HTML]{FFEB84}0.03671 \\
P & Atan transformed pressure & Sense-HAT & all-in-1 & None & \cellcolor[HTML]{FCB479}0.01005 & \cellcolor[HTML]{FEE382}0.92438 & \cellcolor[HTML]{FDEB84}0.80709 & \cellcolor[HTML]{FFEB84}0.15813 & \cellcolor[HTML]{FEEB84}0.04238 & \cellcolor[HTML]{FBAA77}0.01005 & \cellcolor[HTML]{FEE683}0.92627 & \cellcolor[HTML]{FDEB84}0.856712 & \cellcolor[HTML]{FBA877}0.07473 & \cellcolor[HTML]{FFEB84}0.03428 \\
HI & Atan transformed heat index & Sense-HAT & all-in-1 & None & \cellcolor[HTML]{FFEB84}0.02183 & \cellcolor[HTML]{FEE783}0.94771 & \cellcolor[HTML]{FDEB84}0.82037 & \cellcolor[HTML]{FEE983}0.15561 & \cellcolor[HTML]{FDEB84}0.059 & \cellcolor[HTML]{FFEB84}0.02164 & \cellcolor[HTML]{FEE883}0.9383 & \cellcolor[HTML]{FDEB84}0.84311 & \cellcolor[HTML]{FBB078}0.08332 & \cellcolor[HTML]{FDEB84}0.07736 \\
H & humidity scaled 0-1 & Sense-HAT & all-in-1 & None & \cellcolor[HTML]{FFEB84}0.073 & \cellcolor[HTML]{F8696B}0.21 & \cellcolor[HTML]{FBA376}0.1352 & \cellcolor[HTML]{F8756D}0.0262 & \cellcolor[HTML]{FFEB84}0.0252 & \cellcolor[HTML]{FFEB84}0.053 & \cellcolor[HTML]{F8696B}0.188 & \cellcolor[HTML]{FA9C74}0.121084 & \cellcolor[HTML]{F97B6E}0.03011 & \cellcolor[HTML]{FFEB84}0.03113 \\
\textbf{p} & \textbf{Probability of BLE devices} & \textbf{All} & \textbf{ble\_devices} & None & \cellcolor[HTML]{FFEB84}\textbf{0.13118} & \cellcolor[HTML]{FFEB84}\textbf{0.98806} & \cellcolor[HTML]{FFEB84}\textbf{0.38963} & \cellcolor[HTML]{FFEB84}\textbf{0.17093} & \cellcolor[HTML]{F9EA84}\textbf{0.12864} & \cellcolor[HTML]{FFEB84}\textbf{0.13118} & \cellcolor[HTML]{FFEB84}\textbf{0.99996} & \cellcolor[HTML]{FFEB84}\textbf{0.494271} & \cellcolor[HTML]{F9EA84}\textbf{0.26898} & \cellcolor[HTML]{F4E884}\textbf{0.25697} \\
\textbf{p} & \textbf{Probability of WiFi devices} & \textbf{All} & \textbf{wifi\_devices} & None & \cellcolor[HTML]{F8696B}\textbf{0} & \cellcolor[HTML]{FFEB84}\textbf{0.99999} & \cellcolor[HTML]{FFEB84}\textbf{0.37118} & \cellcolor[HTML]{FDEB84}\textbf{0.20366} & \cellcolor[HTML]{F7E984}\textbf{0.15017} & \cellcolor[HTML]{F8696B}\textbf{0} & \cellcolor[HTML]{FFEB84}\textbf{1} & \cellcolor[HTML]{FEEB84}\textbf{0.655721} & \cellcolor[HTML]{FAEA84}\textbf{0.23514} & \cellcolor[HTML]{F3E884}\textbf{0.27778}
\end{tabular}}
\end{minipage}
\end{center}
\end{sidewaystable}
\subsubsection{Single Data Streams}
\begin{figure}
    \includegraphics[scale=.28]{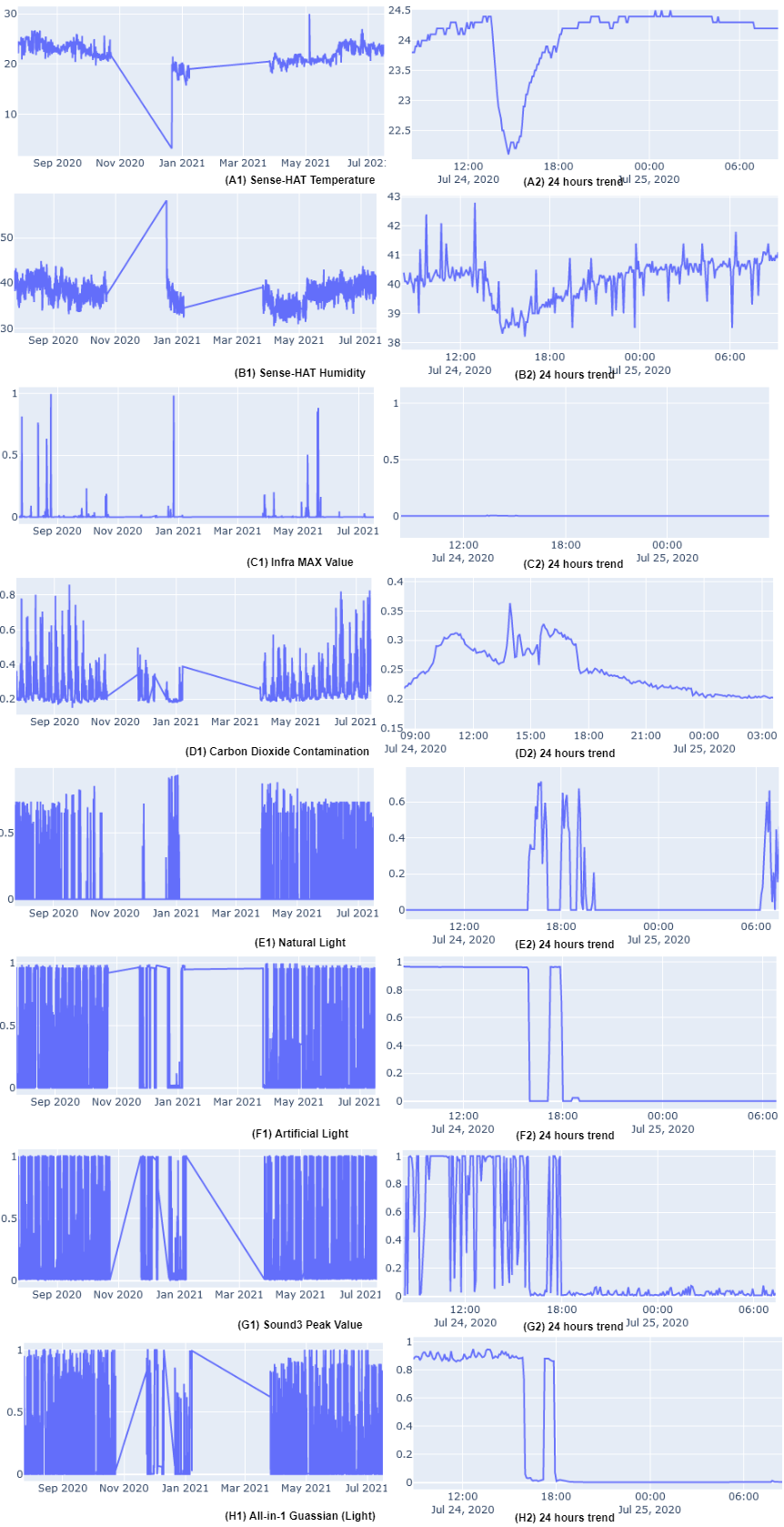}
    \caption{Single Data Streams}
    \label{fig:SingleDataStreams}
\end{figure}

Figures \ref{fig:SingleDataStreams}hold visualization of some of the unique data streams. We structured Sub-figures as a 1x2 matrix where the left side (x1) graph shows all data and the right side (x2) graph shows one-day activity.
The left side graph of figures \ref{fig:SingleDataStreams}(A1) and \ref{fig:SingleDataStreams}(B1) that there is a sudden dip in temperature and increased humidity near the end of October 2020 till the end of December 2020. We also observe that Air Quality is dropping abruptly at the same time. Though these events resulted from disconnection and/or power failure on the device, both were considered anomalous and kept in the data sets; we will discuss other aspects later in the paper.
In Figures \ref{fig:SingleDataStreams}(E2) and \ref{fig:SingleDataStreams}(F2), we observed that the 24 hours trend of artificial light, and natural is identical except a few activities of artificial light can be found in the nighttime. The light sensor in the all-in-1 device, figures \ref{fig:SingleDataStreams}(H1) and (H2), share similar trends. It is noticeable that natural light trends are gradual compared to artificial light.
We also noticed that activities related to Sound, Light, CO\textsubscript{2}, infra, BLE devices and particulate concentration are stable and low-valued at night time. Thus we decided to filter data sets based on daylight conditions as well. We also observe a regular (not everyday) activity before the start of daylight time; this issue has consequences which will be discussed later in the paper.
\subsubsection{Multi Data Streams}
\begin{figure}
    \includegraphics[scale=.28]{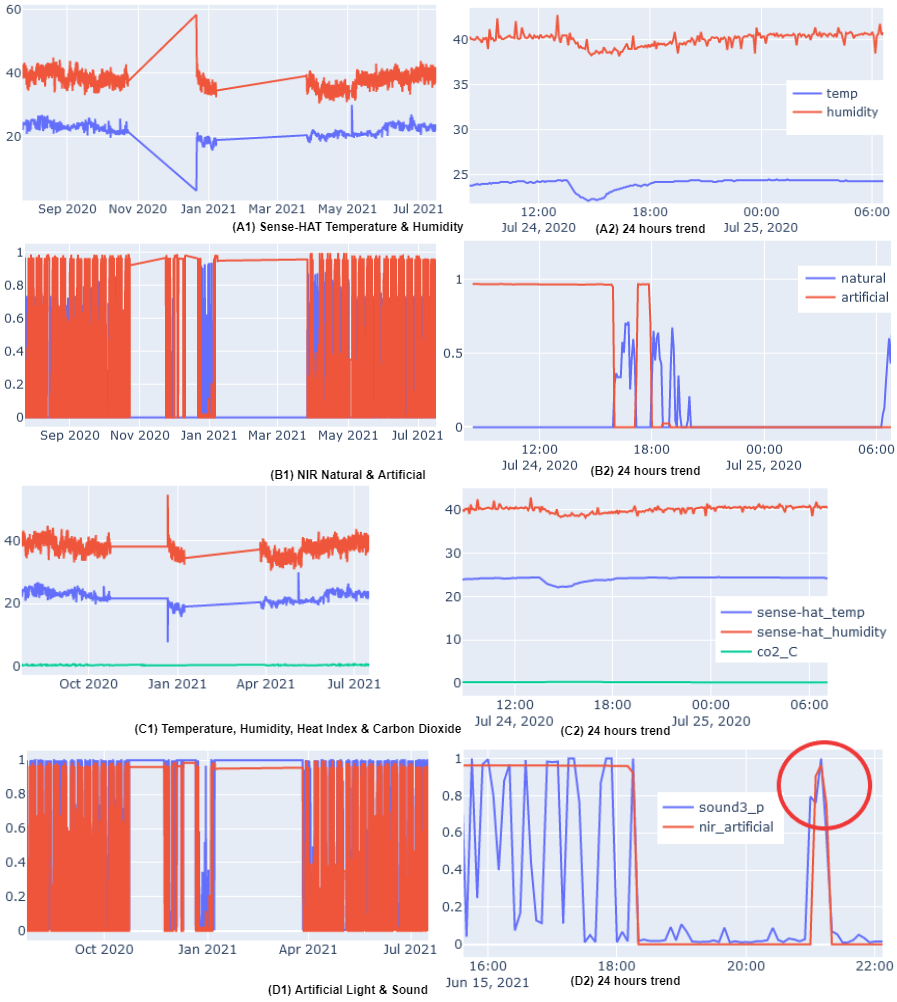}
    \caption{Multi Data Streams}
    \label{fig:MultiDataStreams}
\end{figure}

Analysing relations between different data streams is difficult, ineffective and time-consuming when done separately. So we visualised multiple data streams to analyse the relations demonstrated in figure \ref{fig:MultiDataStreams}. For example, in figure \ref{fig:MultiDataStreams}(A1), it can straightforwardly be noticed that the values of temperature and humidity go opposite directions around the end of October 2020 till the end of December 2020. We can also notice the relation between natural and artificial light in Figures \ref{fig:MultiDataStreams}(B1) and (B2). There are two possible types of multi-data streams in the given setup, intra-device and inter-device. Visualising multiple data streams from one device is comparatively easy as there are a limited number of combinations. On the other hand, inter-device data stream combinations can be enormous, so we chose only (14) unique sensors' data streams, see \textbf{bold} items in Table \ref{tab:DataStreamsDetails}. We choose a couple of inter-device combination graphs for demonstration which can be seen in Figures \ref{fig:MultiDataStreams}(C1) and \ref{fig:MultiDataStreams}(D1). Figure \ref{fig:MultiDataStreams}(D2) has a different situation plotted in which a fire alarm went off at night time and visited by a staff member to evaluate the situation, which triggered the light in the room as seen in the red circle. This activity is perfect to be considered a contextual anomaly. From the left side graph, we can see a regular activity of sound and light in the daytime. Later in this paper, we will evaluate ML models by considering two things i) the regular activity detected as an anomaly, and ii) the sound and light activity around 2100 hours is considered an anomaly.
\subsection{Data Scaling and Reduction Techniques}
The machine learns from the provided data instead of legacy statistical or mathematical algorithms in the ML context. It makes pre-processing of data sets an essential part of the process. Data standardisation is being largely practised for pre-processing data sets before performing ML. It drastically decreases the size of the input sample (in some cases) and time for a model generation compared to non-scaled data. We adopted two techniques for standardisation, StandardScaler and MinMaxScaler.
Standardisation techniques can only convert data into a certain range and can be reversed but can not reduce the dimensions of the input sample in the case of multivariate data. So, we used reduction techniques to convert multivariate data into uni-variate. Reduction techniques help in reducing ML model generation time to a minimum. The resulting data sample from reduction techniques is computationally expensive to reverse. Which makes it hide properties of individual data streams or sensor values, e.g. value of temperature and humidity can only be known by the edge device but will be kept unknown by the fog or cloud device. Scaling techniques are feasible on cloud/fog where a complete data set is available to evaluate a given ML model. We did not consider data scaling for ML models destined to run on edge devices (microcontrollers).
We added another dimension to data sets after applying pre-processing techniques to convert the data into time series, and the resulting sample was three-dimensional.
We used two scaling techniques and five reduction techniques on the available data to evaluate the time difference for model generation. We experienced that scaling techniques take less time (a few microseconds) versus reduction techniques which takes 1500 to 2127 microseconds to execute the process.
\subsubsection{Scaling Techniques}
We used the following data scaling techniques for this work.
\textbf{Standard Scaler} calculate the mean and standard deviation of the input sample before applying equation \ref{eq:standardscaler}.
In equation \ref{eq:standardscaler} \textit{SSd} is the standard scaler output sample of input sample \textit{d}, \textit{u} is equal to the mean of sample \textit{d} and \textit{s} is equal to the standard deviation of input sample \textit{d}.
    \begin{equation}
        \label{eq:standardscaler}
        SSd=\frac{(d-u)}{s}
    \end{equation}
The resulting output sample has a mean=0 and standard deviation=1. We used the StandardScaler function from the sklearn library to perform this scaling operation.\\
\textbf{MinMax Scaler} is simpler than StandardScaler, there is no pre-calculation required as compared to StandardScaler, and most frequently used for input sample standardisation. The output sample is in the range of 0 to 1. The corresponding output value of the minimum value in the sample will be 0, and the corresponding output value of the maximum value in the sample will be 1. These values are calculated using the equation \ref{eq:minmaxscaler}. We used the MinMaxScaler function from the sklearn library to perform this scaling operation.
    \begin{equation}
        \label{eq:minmaxscaler}
        MMd=\frac{(\mathrm{d}-\mathrm{d}_{min})}{(d_{max}-d_{min})}
    \end{equation}
In equation \ref{eq:minmaxscaler} \textit{MMd} is the MinMax scaler output sample of input \textit{d}, d(min) is the minimum value in input sample \textit{d} and \textit{d(max)} is the maximum value in input sample \textit{d}.
\subsubsection{Reduction Techniques}
We used the following data reduction techniques for this paper.\\
\textbf{Average} is the sum of all values divided by the number of values resulting in a single value for each sample. Average can reflect the central tendency of multiple data streams while converting the input sample into univariate. Average requires the least processing resources as compared to other pre-processing techniques. We used the average function from the NumPy library to execute this operation on the multi-variate input samples.
\begin{equation}
        \label{eq:mean}
        \bar{m} = (\frac{1}{n})\sum_{i=1}^{n}x_i
    \end{equation}
\textbf{Standard Deviation (SD)} results in a univariate data stream that can reflect the spread of a multivariate input sample. It takes slightly more processing resources than average as the average input sample is a prerequisite for the SD equation to be executed. We used the std function from the NumPy library to execute this operation on multi-variate input samples.
\begin{equation}
        \label{eq:std}
        \sigma = \sqrt\frac{\sum_{i=1}^n(x_i-\bar{x})^2}{n}
    \end{equation}
\textbf{Median Absolute Deviation (MAD)} calculates variability in the input sample, it is more computationally complex than SD because it is dependent on the median value of the input sample. MAD is more resilient in terms of outlier detection as compared to SD. We used the median\_abs\_deviation function from scipy.stats library for this operation.
\begin{equation}
        \label{eq:mad}
        MAD =median(x_i - \bar{x})
    \end{equation}
\textbf{Kurtosis (Ku)} calculates the relative peakedness of an input sample, it requires both average and SD of the input sample thus the computational power requires is more than the previous techniques. We noticed that Ku is effective on larger data points in terms of influencing anomaly detection. We used stats.kurtosis function from scipy library for this operation.
    \begin{equation}
        \label{eq:kurtosis}
        \mathrm{K} = \frac{1}{n}\sum_{i=1}^{n}
        \frac{(x_i-\bar{x})^4}{\sigma^4}    
    \end{equation}
\textbf{Skewness (Skew)} calculates the trends of the input sample, it can be a normal, negative or positive skewness value. Skew is the most computationally complex in our discussed techniques, it requires precomputed average and SD of the input sample. It is also effective on larger data points where a curve can be formed. We used stats.skew function from scipy library for this operation.
\begin{equation}
        \label{eq:skew}
        \mathrm{Sk} = \frac{1}{n}\sum_{i=1}^{n}
        \frac{(x_i-\bar{x})^3}{\sigma^3} 
    \end{equation}
\subsection{Data Conversion to Time Series}
We tried and compared different algorithms to convert series data in a time-series format, i.e. each row contains the number of future rows. In streaming data scenarios, anomalies are categorised based on data trends instead of points, e.g. the temperature in daytime hits 30\textsuperscript{o}C. In contrast, at night time, it remains below 18\textsuperscript{o}C. Considering a microcontroller without an internal clock can only be aware of the context be current values rather than time. The ML model shall be trained using a time-series-based input sample to achieve this functionally. Let us say the dimensions of the input sample are [Rows, data points], e.g. [36484, 14], dimensions of the resulting sample become [Rows, Time Steps, data points], e.g. [36484, 74, 14]. Let us say R represents data rows in the data set, T represents the number of required time-steps for each sample, X represents the use-able rows, and Y is the resulting time-series sample.
\begin{equation}
\begin{aligned}
X \in \{R0, R1, R2, \dots, R-T \} \\
Y \in \{X+1, X+2, X+3 \dots, X+T \}
\end{aligned}
\end{equation}
\subsection{Anomaly Detection Techniques Selection}
We used the following anomaly detection techniques in this paper.
\subsubsection{OneClassSVM (OSSVM)}
Support Vector Machine (SVM) is one of the most common ML methods \cite{Djenouri2019}. SVM is primarily used for classification (supervised ML) but can also be adopted for clustering (unsupervised ML). SVM is memory efficient, flexible, and suitable in high dimensional spaces and even works with a smaller number of samples compared to dimensions. It has a sub-method, OneClass for outlier-detection, that tries to discover decision boundaries to achieve maximum distance between data points and source by using a clustering mechanism. The main idea behind OneClass was stalled because of its incompetence in finding outliers and determining non-linear decision boundaries. However, with the introduction of soft margins and kernels, these issues were resolved \cite{Amer2013}. OneClass SVM splits all given data points from the source and amplifies the distance from this subspace to the source in the training phase. The function returns a binary output for each input row where \textit{+1} means smaller distance and \textit{-1} means larger distance where larger distance considers an anomaly \cite{Scholkopf2000}. It is widely used in various applications for both supervised and unsupervised learning methods. It is also heavily adopted in academia. An anomaly classifier using SVM was proposed \cite{Araya2017} for detecting abnormal consumption behaviour. A method proposed by \cite{Ferdoash2015} to calculate excessive airflow in Heating Ventilation and Air Conditioning (HVAC) units in a large-scale Building Management System (BMS). They also calculated the pre-cooling start time for reaching the required temperature using temperature sensors. \cite{Jakkula2011} the proposes OneClass SVM for anomaly detection in smart home environments using publicly available smart environment data sets. \cite{Himeur2021a} proposed a method to detect anomalous power consumption in buildings. OCSVM is highly effective on point anomalies and can be inferred on fog devices to be used in real-time environments.
\subsubsection{Isolation Forest (IF)}
IF is one of the top-most used algorithms in the outlier detection domain because of its speed and simplicity. IF is based on ensemble learning. The idea behind IF is that randomly developed decision trees can quickly isolate an outlier in the data set instead of detecting outliers using density or distance from other samples. Outliers are isolated because of the shorter path in the tree as they have fewer relations with other data points \cite{Liu2008}. In terms of functional performance in outlier detection, IF is the most popular algorithm \cite{Buschjager2020}. We use the IsolationForest function from the SKLearn library to perform model generation. The function requires all samples as input and return a list of anomaly score for each sample. IF is also effective for point anomalies only. It is not suitable for fog devices in real-time scenarios as it requires a complete dataset.
\subsubsection{CNN}
In Deep Neural Networks (DNN), Convolutional Neural Network (CNN) is on the most wanted neural networks list. The name "Convolutional" comes from the matrixes-based linear operation. CNN models consist of multiple layers, e.g. max-pooling, fully-connected, and others \cite{Albawi2018}. It brings significant improvement in computer vision (CV), Time series prediction and Natural Language Processing (NLP). It covers a wide range of application scenarios by providing single and multidimensional layers, i.e. 1-D CNN, supporting Time Series Prediction and Signal Identification. 2-D CNN enables Image Classification, Object Detection, Image Segmentation and Face Recognition and 3-D CNN, which helps in Human Action Recognition and Object Recognition/Detection \cite{Li2021}. In contrast with other classification approaches, e.g. feature-based, CNN can find and learn relations and generate in-depth features from time-series data streams automatically, e.g. speech recognition, ECG, price stocks, pattern recognition, rule discovery, and many more \cite{Zhao2017}. All platforms support CNN, i.e. Edge (microcontrollers), Fog (RaspberryPi, Mobile Platforms) and Cloud (High-performance Linux, Windows or Other OSes). We implemented CNN by using TensorFlow API. 
\subsubsection{RNN}
A recurrent Neural Network (RNN) is also a type of DNN, and it is designed with built-in memory, making it more suitable for time-series-based data streams. Another feature of RNN is that it can process information in bi-directional instead of forwarding direction only. Typical RNN has a known issue of vanishing or exploding gradient, which affects its accuracy and overall performance. With the help of Long Short-Term Memory (LSTM) \cite{Hochreiter1997}, which is designed with a memory cell to hold information over a period of time, this problem can be resolved. LSTM is complex but sophisticated, and has three gates input, output and forget. RNN models can predict the future value from time-based input compared with the data sample to calculate the loss. If the loss is greater than the threshold (pre-computed using the training sample), the data sample can be categorised as an anomaly. LSTM is widely used in various applications commonly based on time-series data. LSTM is available only on Fog and Cloud devices using the TensorFlow library. Anomaly detection in a time-series context is a significant application of LSTM.
\section{Experimentation Results}
This section will discuss the results of different combinations of data pre-processing and ML models. We tested selective TF models on all platforms (Cloud-Fog-Edge) and  SKLearn  models on Cloud and Fog only.  SKLearn  models predictions are binary (Anomaly=-1, Normal=1) whereas TF models were based on future prediction, so the output was non-binary. Results for TF models were calculated using a two steps process. First, we calculated the Mean Absolute Error (MAE) for the predicted loss method using equation \ref{eq:mae} and threshold by using equation \ref{eq:threshold}.

\begin{equation}
        \label{eq:mae}
        MAE = \frac{1}{n}\sum_{t=1}^{n}|y-x|
    \end{equation}
The equatref{eq:mae} calculates the mean absolute error (average loss) of all input samples by calculating absolute loss for each sample, where n represents a number of samples, y represents predicted and x represents expected values of each sample.
\begin{equation}
\begin{aligned}
\label{eq:threshold}
Threshold = (8\cdot \sigma(MAE)) + MAE \\
\sigma -> Standard Deviation
\end{aligned}
\end{equation}
Equation \ref{eq:threshold} dynamically calculates the threshold by calculating the standard deviation of MAE, multiplying it eight times and adding it with MAE. If the resulting loss of an input sample is greater than the threshold, the sample is considered anomalous.
\subsection{Architectural Configurations}
As discussed previously, we are using four types of ML Models to train and test available data sets. These models are from two different APIs, Sci-kit Learn (SKLearn) and TensorFlow (TF). SKLearn and RNN based models are available on Cloud and Fog platforms, whereas CNN is also deployable on edge devices. In this section, we will discuss the configurations of each algorithm. We configure the OCSVM model with 0.5 nu, "auto" gamma and "RBF" kernel parameters. We configure IF model for "auto" contamination parameter. Early-Stopping to monitor loss with min\_delta=1e-2 and patience=3 was configured for both CNN and RNN models. We converted the dataset for both NN models into 74-time steps. We also fixed 100 epochs (max), adam optimizer, and batch size to be 10 for both NN models. Our CNN model requires TensorFlow version 2.1.1 and RNN on the 2.4.1 version. We configured CNN models with Conv1D layer, kernel size of 32, 5 filters and mean-squared error for loss calculation. We used LSTM layers for RNN models with 32 neurons and mean-absolute-error for loss calculation.

\subsection{Data Streams' Configurations}
\label{sec:DataStreamsConfigurations}
We divided our data sets into two sub-datasets depending on daylight conditions, e.g., day time sub-dataset (DT) and night time sub-dataset (NT). We used unconditional data set (UC) for ML models as well. We implemented these scenarios on these two types of streams. Converting datasets into sub-datasets reduces the ML model generation time as well as inference time. It also supports (in some cases) the implementation of point-based anomaly detection, e.g. illumination. Events at nighttime can be detected with high accuracy and low computational resources if the ML model is trained using the NT sub-dataset. On the other hand, sub-datasets are limited to specific circumstances only, e.g. if the buildings are designed to be illuminated 24x7.

i.	Univariate (Single Data Streams): each data stream from all devices was used to train, test, and analyse models. Because these Data Streams were already uni-variate, reduction techniques were not applicable.
ii.	Multivariate (Multiple Data Stream): There can be enormous possible combinations between intra-device and inter-device data streams. Research has already been conducted about relations between physical channels like temperature-humidity with CO\textsubscript{2}   \cite{Liu2017}. Showing all possible combinations of multi-data streams is overwhelming; thus, we have presented results of a few of these combinations and preserved all models and results stored for detailed analysis.
\subsection{Results}\label{Results}

\subsubsection{Univariate vs Multivariate} 
\begin{figure}
    \includegraphics[scale=.5]{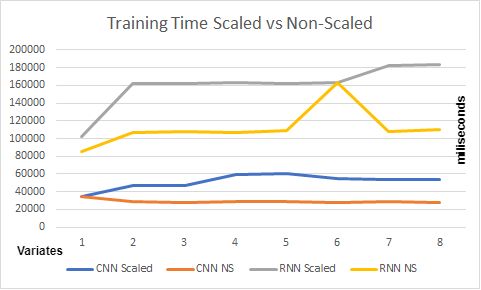}
    \caption{Scaled vs Non-Scaled and RNN vs CNN Model Training Times}
    \label{fig:TrainingTime}
\end{figure}

Reduction techniques returns univariate data so the model training time is identical for all number of data stream combinations. Total training time also depends on the number of epochs executed before early stopping condition becomes true. Figure \ref{fig:TrainingTime} shows model training times of scaled vs non-scaled dataset, it can be observed that scaled dataset took more time for training in both CNN and RNN methods. It is also obvious to see that RNN CNN is efficient when compared to RNN. Due to limited knowledge of known anomalies in the dataset, it is difficult to determine overall efficiency of ML models. 
\subsubsection{Detecting Anomalies using Individual Sensor Data Streams (Univariate)}
\begin{figure}
    \includegraphics[scale=.45]{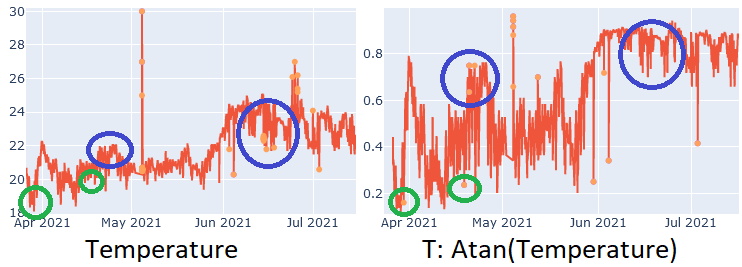}
    \caption{Temperature vs Atan (Temperature) Comparison}
    \label{fig:sense-hat_temp-t_comparision}  
\end{figure}

\begin{figure}
    \includegraphics[scale=.45]{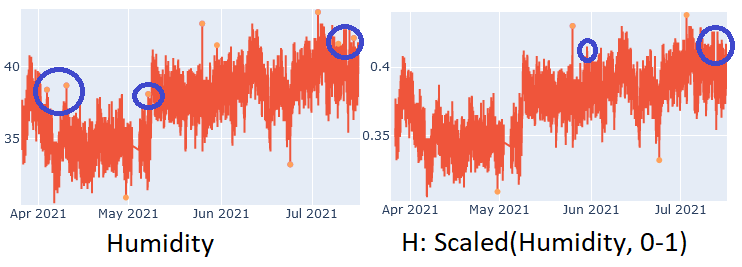}
    \caption{Humidity vs Percentage (Humidity) Comparison}
    \label{fig:sense-hat_humidity-H_comparision}
\end{figure}

A comparison of temperature with edge-processed T data streams, which is atan (temperature), from the sense-hat device. We had 32 data streams, out of which 14 were from unique sensors, and 18 were associated streams. While comparing different sensor and associated data streams, we found that atan converted data streams required a lesser threshold value to find anomalies in novel data. The transformed data streams were ineffective at certain stages where change suddenly fluctuated. As seen in circled in blue colour where anomalies are shown in orange dots in figure \ref{fig:sense-hat_temp-t_comparision}, a few anomalies found in T, all at a lower temperature, was not detected in the temperature model can be seen in green circles. When it comes to humidity, the edge-processed scaled data stream H was less sensitive as compared to the unprocessed data stream, as demonstrated in figure \ref{fig:sense-hat_humidity-H_comparision}, the blue circles highlight the difference. Since we generated models for three environmental conditions, we found that the sum of anomalies found in two daylight condition-based data sets (dark=0, light=1) was equal to the number of anomalies found in the unconditional data set.

We also noticed that there is no difference in non-scaled streams vs scaled streams in temperature and its associated data streams, e.g. T. Whereas other sensors and associated data streams show different results, e.g. a number of anomalies found original data stream of humidity sensor were noticeably different from StandardScaler but comparatively similar with MinMax. We observed that StandardScaler decreases sensitivity resulting in lesser anomalies as compared to the non-scaled data stream. It was also observed that MinMaxScaler increased sensitivity resulting in more anomalies. 
We found an obvious difference when comparing a number of anomalies in pressure (P) and particulate concentration (M) data streams where StandardScaler results in drastically increased sensitivity, the number of anomalies are greater using a smaller threshold level.
On the other hand, anomalies found in Carbon dioxide (CO\textsubscript{2}) in scaled versions of data streams were fewer as compared to non-scaled data stream based models, which point toward a decrease in sensitivity. Another noticeable trend in the number of anomalies is that the sum of both conditional anomalies was marginally greater than the unconditional data set except for standard scaler based models. We found a unique trend in artificial sensor condition-based models. No anomalies were found in non-scaled and MinMax scaler models in conditional data sets, but standard scaled models found anomalies. Anomalies found in unconditional data set based models were similar to non-scaled and scaled models. Sound sensor-based models show an opposite reaction when it comes to anomalies; we found zero anomalies in UC and DT. Whereas NT based models found anomalies, non-scaled and MinMaxScaler models were pretty much similar. However, the StandardScaler model found more anomalies that represent increased sensitivity similar to previously discussed pressure and particulate concentration models.
\subsubsection{Detecting Anomalies using Intra-Device (Multivariate)}
The total number of unique intra-device combinations of data streams was 626. We choose a few of them for analysis in this paper. We noticed that most of the data preprocessing techniques could find almost similar anomalies in the sense-hat device (all data streams), except MinMaxScaler, which was extremely sensitive, and MAD was too insensitive. Kurtosis and Skewness were not effective. Zero anomalies were found when implemented on the temperature and humidity (Temp-Humidity) set. The behaviour of MinMaxScaler was the same in Temp-Humidity but turns regular when used on all other associated streams, i.e., T, P, H and HI (T-P-H-HI) MAD were also able to find the same contextual anomalies on this set. When looking at the results of all data streams in All-in-1, we found that MAD was most sensitive on UC and most insensitive on DT (zero anomalies). The average was not effective (a few anomalies detected) on NT and UC, whereas it could find the same contextual anomalies as other techniques. We noticed that temperature sensor readings were regularly dipping randomly and abruptly, which was one of the reasons for its influence over other data streams and thus on statistical outcomes. Looking at other models in all-in-1 devices, excluding temperature-related values, we found few anomalous activities.
\subsubsection{Detecting Anomalies using Inter-Device Multiple Data Streams
(Multivariate)}
\begin{figure}[ht]
    \centering
    \includegraphics[width=0.5\textwidth]{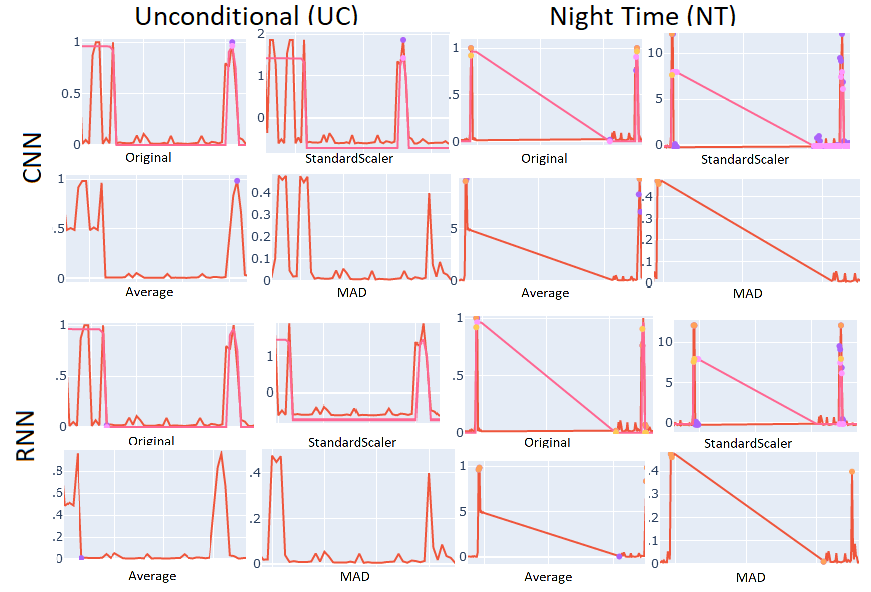}
    \caption{Sound \& Light Known "Anomalous Activity" Analysis}
    \label{fig:sound_light_combination_comparison}
\end{figure}

As discussed in an earlier section about the one known anomalous activity based on sound and light sensors' data, we analyzed the particular activity to learn the effectiveness of different algorithms and pre-processing techniques. We found that the CNN model with scaled, non-scaled and average sound and artificial values can spot the anomalous activity without spotting false positives (usual everyday activity). In contrast, RNN models were not successful in detecting the particular activity, as shown in figure \ref{fig:sound_light_combination_comparison}. We also noticed that false positives were found in all models, along with detecting anomalous activity in the NT dataset. We also found that SKLearn based models overwhelmed false positives in all datasets.
\subsubsection{Point, Contextual, Combined Anomalies}
\begin{figure}[ht]
    \centering
    \includegraphics[width=0.5\textwidth]{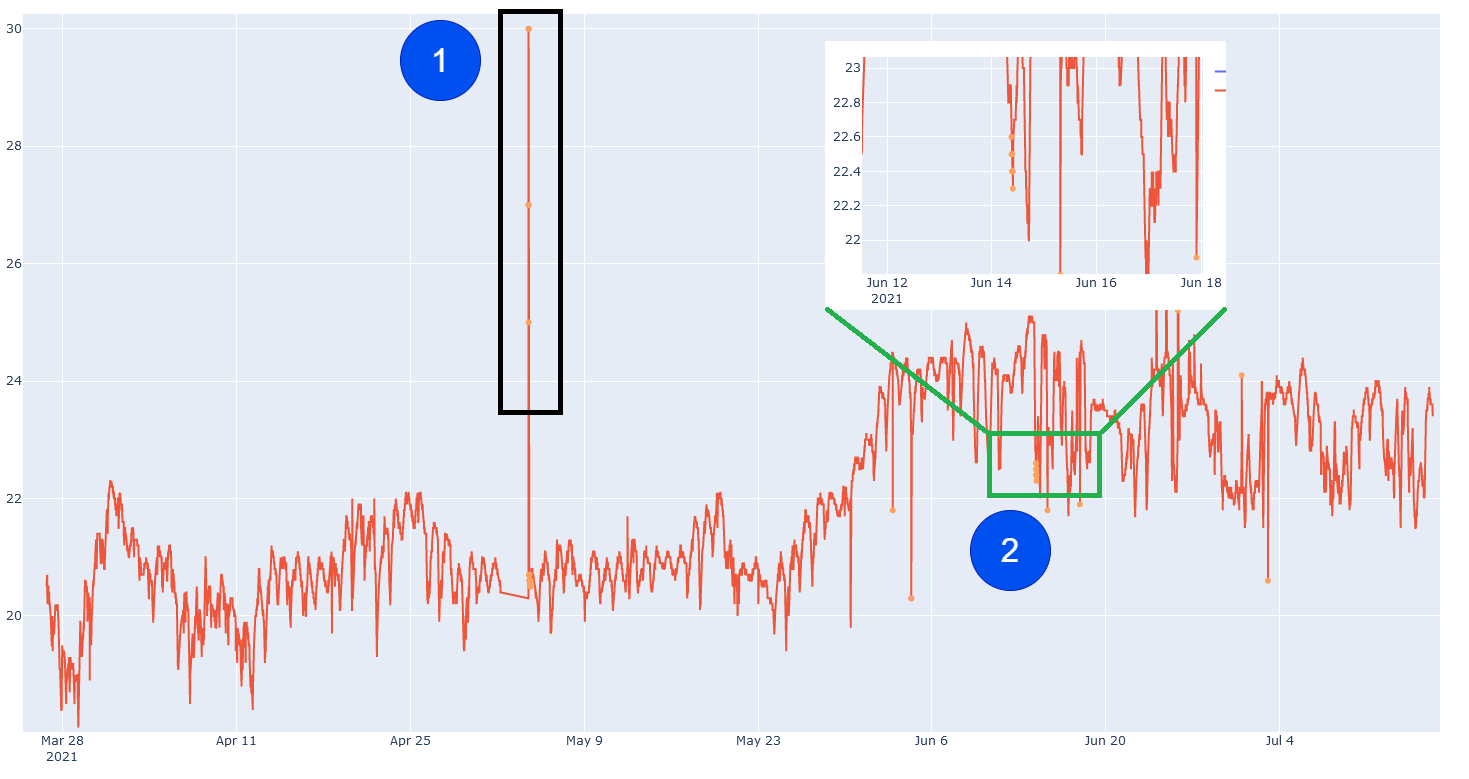}
    \caption{1-Point Anomaly vs 2-Contextual Anomaly in Temperature Data Stream}
    \label{fig:point_vs_contextual}
\end{figure}

\begin{figure}[ht]
    \centering
    \includegraphics[width=0.5\textwidth]{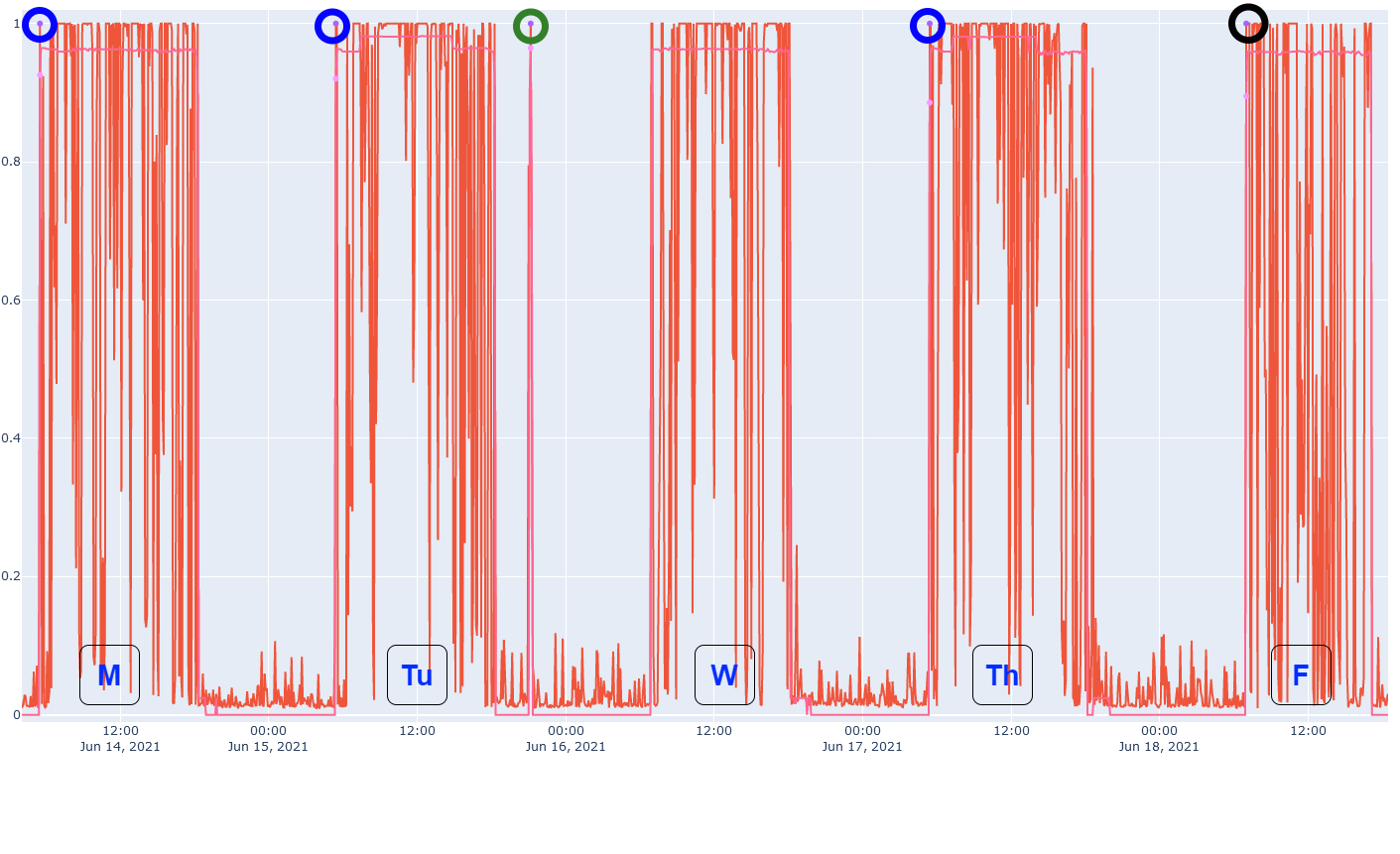}
    \caption{Combined Contextual Anomalies in Sound and Artificial Light Data Streams}
    \label{fig:combinedAnomaly}
\end{figure}

Looking closely at figure \ref{fig:point_vs_contextual}, the two highlighted portions of the timeline of the temperature data stream from the sense-hat device. We observed at the end of April 2021 temperature sensor malfunctioned, resulting in an extreme increase to 30\textsuperscript{o}C. Another event marked anomalous in highlighted point 2 shows a sudden dip in temperature from 22.6\textsuperscript{o}C to 22.9\textsuperscript{o}C detected. While looking at historical data, both points are in the normal range, but this activity is considered anomalous in context. Figure \ref{fig:combinedAnomaly} shows the combined activity of artificial light and sound for the week commencing on June 14, 2021. In the context, office activity started early, i.e. at 0530 hours on Monday, Tuesday, and Thursday and was detected as anomalous True Positive (TP). The office starts at 0700 hours on Friday and Wednesday, as shown in the black circle. The Friday morning activity was detected as False Positive (FP). On the other hand, the Wednesday activity was accurately detected as True Negative (TN). In addition to day start activities, a TP anomaly was detected around 2100 hours due to a response initiated as a result of a (separately operated) fire alarm. 
\section{Related Work}
There are some suggestions for supervised anomaly detection methods\cite{Liu2015} \cite{Laptev2015}. The results are promising, but labelled data is rare in the real world. Perhaps unsupervised ML methods have become the focus of attention because of the excellent performance and the flexibility provided \cite{Li2021a}. The scope of anomaly detection is not limited to specific areas. However, everywhere e.g., industry \cite{Oh2018}, financial systems \cite{Gran2010}, healthcare and maintenance of spacecraft by detecting anomalies \cite{Gupta2014a}, cyber-physical system \cite{Luo2021}, and smart buildings \cite{Araya2016}.
\subsection{Anomaly Detection Techniques for IoT Data}
Research conducted by Microsoft \cite{Ren2019b} led to the development of an algorithm for detecting anomalies in time-series data using residual spectrum processing and convolutional neural networks (SR-CNN). However, they were mainly concerned about stationery and seasonal data, resulting in ineffective results on non-stationary data. Data from Surface-mounted audio sensors used with semi-supervised CNN auto-encoders \cite{Oh2018} to detect faults in industrial machinery. A deep autoencoders based model has been proposed for detecting spectrum anomalies in wireless communications \cite{Feng2017a}. The model developed in this work is to detect anomalies that may occur due to an abrupt change in the signal-to-noise ratio (SNR) of the monitored communications channel. In a critical infrastructure environment, if phasor data is manipulated, the control centres may take the wrong actions, negatively impacting power transmission reliability. To mitigate this threat \cite{Yan2015} proposed a deep autoencoder technique. The \cite{Zhang2018c} study uses data from a number of heterogeneous IIoT sensors, including temperature, pressure, vibration, and others, to develop an RNN-LSTM based regression model to predict failures in pumps at a power station. A new RNN-LSTM based method was developed \cite{Hundman2018a} to detect anomalies in a massive amount of telemetry data from spacecraft. They also offered a method for evaluating that was non-parametric, dynamic, and unsupervised. Another solution proposed \cite{Wu2020a} to detect anomalies in multi-seasonality time-series data using RNN-GRU also proposed a Local Trend Inconsistency metric on top of their proposed anomaly detection algorithm. The authors of \cite{Marti2015a} proposed a combination of Yet Another Segmentation Algorithm (YASA) and OneClassSVM (OCSVM) in order to detect anomalous activities in turbomachines in the petroleum industry. The authors of \cite{Aurino2014} used OCSVM to detect gunshots from audio signals. OCSVM grouped with DNN used to detect road traffic activities by \cite{Rovetta2020}. Isolation Forest (IF) was used to detect anomalies in smart audio sensors \cite{Antonini2018}. IF is also used, in combination with order-preserving hashing techniques, to detect anomalies by \cite{Xiang2020}. Another novel approach proposed by \cite{Farzad2020a} uses autoencoder based IF for log-based anomaly detection.
\subsection{Environmental Monitoring within Buildings}
In today's world, human beings spend 90\% of their time in built environments which includes residential, commercial, education, as well as transport, i.e. vehicles, \cite{brady2021rating}.  Monitoring an indoor environment is different from industrial or mission-critical infrastructure, where normal activities are largely known because of the heterogeneous nature of activities. There are several environmental monitoring applications other than anomaly detection, e.g. Energy Monitoring, Comfort Level Monitoring. Environment monitoring is well researched. The heterogeneous nature of environments requires the selection of the suitable parameters, sensors technologies, communication mediums, placement and power arrangements. Major parameters in this domain are temperature, humidity, carbon emissions, illumination, airflow, and occupancy \cite{Hayat2019}. Air Quality (AQ) is becoming a critical matter. WHO reported that there are almost 7 million premature deaths are being caused by air pollution annually \cite{WHO2021}. Authors of \cite{Saini2020} presented a survey of system architectures used for Indoor Air Quality (IAQ) data collection as well as methods and applications for prediction. Indoor environment quality plays an essential role in the health and well-being of human beings, \cite{Clements2019} presented a living lab to simulate real office spaces to support further research on environmental monitoring in the built environment. Occupancy monitoring is essential to determine air-conditioning and illumination requirements in buildings, \cite{Erickson2014} proposed a wireless sensor network based occupancy model to be integrated with buildings conditioning systems. Based on two seasons of monitoring IAQ and thermal comforts in school building \cite{Asif2020} recorded more than 50\% increase in CO\textsubscript{2} levels during class times. Thermal comfort has critical importance for the well-being and productivity of occupants in indoor environments, \cite{Valinejadshoubi2021} proposed an integrated sensor-based thermal comport monitoring system for buildings which also provides the virtual visualization of thermal conditions in buildings. Authors of \cite{NgahNasaruddin2019} presented temperature and relative humidity monitoring solutions in high temperature and humid climate environments using well-calibrated thermal micro-climate devices and a single-board microcontroller.
\subsection{Anomaly Detection within Buildings}
Researchers propose a wide variety of methods for anomaly detection in buildings. The diversity of techniques reflects extensive work being done in this domain. Unsupervised learning has been used for fault detection and diagnostics in smart buildings. Authors of \cite{Capozzoli2015} proposed a simple technique based on unsupervised learning that can automatically detect anomalies in energy consumption based on the historically recorded data of active lighting power and total active power. They adopt statistical pattern recognition and ANN along-with other anomaly detection methods. A novel method, Strip, Bind, and Search (SBS), based on unsupervised learning proposed by \cite{Fontugne2013} to help identify devices with anomalous behaviour by looking at inter-device relationships. The authors of \cite{Xu2021} also proposed a data mining based unsupervised learning technique to detect anomalies in HVAC systems; the proposed work also performs dynamic energy performance evaluation. In the models proposed by \cite{Araya2017}, overlapping sliding windows and ensemble anomaly detection were used to identify anomalies. The same authors also proposed a Collective Contextual Anomaly detection using similar techniques in their previous work \cite{Araya2016}. A Generalized Additive Model was proposed by \cite{Ploennigs2013} for diagnosing building problems based on the hierarchy of sub-meters. A Two-Step clustering algorithm based on unsupervised machine learning was proposed by \cite{Poh2020a} to detect anomalous behaviour from physical access data of employees about their job profiles. In a distributed sensor network, an anomaly detection technique was proposed by \cite{Meyn2009} using semi-empirical Markov Models for time-series data. 
In a recent survey conducted by \cite{Himeur2021b}, the authors concluded that anomaly detection techniques could help in the reduction of energy consumption to benefit all stakeholders.
\section{Lessons Learnt and Discussion}
DIY based (single-board computers, microcontrollers, sensors) IoT devices are widely available and becoming easy to deploy. These devices are micro-manageable and cost-effective, but it is a laborious job which leads to various challenges; while doing this research, we learnt the following lessons: (i) missing data due to run-time errors, (ii) threshold calculation, (iii) inter-device synchronisation, (iv) importance of "normal" dataset, (v) an overwhelming number of ML models, (vi) converting time-series data for unsupervised ML processing and (vii) handling interactive graphs.

\textit{Missing data:} DIY devices are prone to configuration, deployment, and handling problems when used for capturing data on a long-term basis. There is no built-in notification system that can alert in case of any error; thus, the errors persist silently for an extended period, ultimately affecting the dataset. During our data-capturing stage, we faced various scenarios where data collection stopped, e.g. device power outage, sensor malfunctions, communication errors, etc. thus; the data is missing during those time slots.

\textit{Threshold calculation:} Anomaly decision in time-series data using an unsupervised approach is based on loss and threshold. The threshold is critical in the decision process and calculating the threshold for each configuration (data stream combinations with sub-datasets). A maximum loss value from a normal dataset (training dataset) can be used as a threshold; to achieve that, an utterly normal dataset (without any capture-time errors) is required.

\textit{Inter-device synchronisation :} Due to multiple device setups, there were synchronisation errors due to missed data in devices at different time slots. Data lost from any single device or frequency differences can result in synchronisation issues. This creates a unique challenge when combining data streams from inter-device. It is recommended to use a single host device for all sensors or create a master table with a single timestamp at the ingester-end to keep data synchronised at capturing stage.

\textit{Importance of "normal" dataset:} For the above-learnt lessons, we observe the critical importance of a completely normal dataset, e.g. without run-time errors (communication, power, hardware).

\textit{An overwhelming number of ML models:} Due to the number of data streams, the number of combinations was in the thousands. The resulting ML models and associated results were overwhelming and difficult to observe and manage. A systematic approach needed to be adopted to handle the heterogeneous configuration of datasets, models, and results.

\textit{Converting time-series data for unsupervised ML processing:} Time-series conversion of data sets using pandas data-frames is far more computationally expensive than using the NumPy library. It is wise to test and compare all available methods for each sub-task before starting mass processing. The result is the same for both methods.

\textit{Handling interactive graphs:} For unsupervised learning approaches for time series, analysing data using interactive graphs is vital but requires extensive computational resources to load and interact graphs with multiple data streams.

\section{Conclusion and Future Work}
In this paper, we captured data streams from various in-situ sensors using different devices with a variety of configurations. We were able to detect point, contextual and combined anomalies. We compared different ML methods combined with several data pre-processing techniques to better understand how to efficiently detect anomalous activities in a smart building environment. We also evaluated the performance of the conditional dataset (based on environmental conditions, e.g. daylight). We found that it can work better for detecting point anomalies as the activities are filtered for certain situations. A clean, anomaly-free dataset is required for model training for better results. Unconventional scaling techniques, e.g., atan, can lower sensitivity for detection and an overhead during the data-capturing process; atan and other conversions can be performed in bulk at any later stage with reasonable computational resources. We explored relations between various sensors in finding anomalies in buildings. We also explored effective techniques to pre-process datasets to optimise ML models. We also introduced an inter-device data synchronisation technique to fill up missing time slots and trim time-series datasets when comparing different datasets. 
Threshold plays a vital role in reducing false positives and increasing true positives. A dynamic threshold calculation is essential to deal with the overwhelming configuration of data streams. The day of the week can also be used as a context for anomaly detection in time-series datasets, but a large dataset is required for modelling. Availability of a dataset with known anomalies will be an important step towards determining overall efficiency of ML models.

\section*{Acknowledgement}
This work is partially supported by EPSRC PETRAS (EP/S035362/1) and GCHQ National Resilience Fellowship.

\bibliographystyle{plainnat}
\bibliography{CleanADSBwithDIYIoT}
\end{document}